\documentclass[runningheads,a4paper]{llncs}

\usepackage{amssymb}
\usepackage{graphicx}
\usepackage{color}
 
\usepackage{url}
\urldef{\mailsa}\path|{alfred.hofmann, ursula.barth, ingrid.haas, frank.holzwarth,|
\urldef{\mailsb}\path|anna.kramer, leonie.kunz, christine.reiss, nicole.sator,|
\urldef{\mailsc}\path|erika.siebert-cole, peter.strasser, lncs}@springer.com|

\begin{document}

\mainmatter  
\pagenumbering{gobble}
\title{View Independent Vehicle Make, Model and Color Recognition Using Convolutional Neural Network}

\author{Afshin Dehghan \hspace{0.25 in} Syed Zain Masood \hspace{0.25 in} Guang Shu \hspace{0.25 in} Enrique G. Ortiz\\
		{\tt\small \{afshindehghan, zainmasood, guangshu, egortiz\}@sighthound.com}}

\institute{Computer Vision Lab, Sighthound Inc., Winter Park, FL}

\maketitle

\begin{abstract}
This paper describes the details of Sighthound's fully automated vehicle make, model and color recognition system. The backbone of our system is a deep convolutional neural network that is not only computationally inexpensive, but also provides state-of-the-art results on several competitive benchmarks. Additionally, our deep network is trained on a large dataset of several million images which are labeled through a semi-automated process. Finally we test our system on several public datasets as well as our own internal test dataset. Our results show that we outperform other methods on all benchmarks by significant margins. Our model is available to developers through the Sighthound Cloud API at \textcolor{blue}{https://www.sighthound.com/products/cloud}
\end{abstract}

\section{Introduction}

Make, model and color recognition (MMCR) of vehicles \cite{SochorCVPR16,Hsieh2014,MVA2013} is of great interest in several applications such as law-enforcement, driver assistance, surveillance and traffic monitoring. This fine-grained visual classification task \cite{compCarDataset,YuICCV2015,XieCVPR2015,KrauseICCV2013,KrauseICPR2014,LinECCV2014} has been traditionally a difficult task for computers. The main challenge is the subtle differences between classes (e.g BMW 3 series and BMW 5 series) compared to some traditional classification tasks such as ImageNet. Recently, there have been efforts to design more accurate algorithms for MMCR such as those in the works of Sochor et al in \cite{SochorCVPR16} and Hsieh et al \cite{Hsieh2014}. Moreover, many researchers have focused on collecting large datasets to facilitate research in this area \cite{compCarDataset}. However, the complexity of current methods and/or the small size of current datasets lead to sub-optimal performance in real world use cases. Thus, there are still considerable short-comings for agencies or commercial entities looking to deploy reliable software for the task of MMCR. In this paper, we present a system that is capable of detecting and tagging the make, model and color of vehicles irrespective of viewing angle with high accuracy. Our model is trained to recognize $59$ different vehicle makes as well as $818$ different models in what we believe is the largest set available for commercial or non commercial use. \footnote{Our system covers almost all popular models in North America.} The contributions of Sighthound's vehicle MMCR system are listed as follows:

\begin{itemize}
\renewcommand{\labelitemi}{\scriptsize$\blacksquare$} 

\item To date, we have collected what we believe to be the largest vehicle dataset, consisting of more than $3$ million images labeled with corresponding vehicle make and model. Additionally, we labeled part of this data with corresponding labels for vehicle color.

\item We propose a semi-automated method for annotating several million vehicle images.

\item We present an end-to-end pipeline, along with a novel deep network, that not only is computationally inexpensive, but also outperforms competitive methods on several benchmarks.

\item We conducted a number of experiments on existing benchmarks and obtained state-of-the-art results on all of them.

\end{itemize}
\section{System Overview}

\begin{figure}
\begin{center}
   \includegraphics[width=\linewidth]{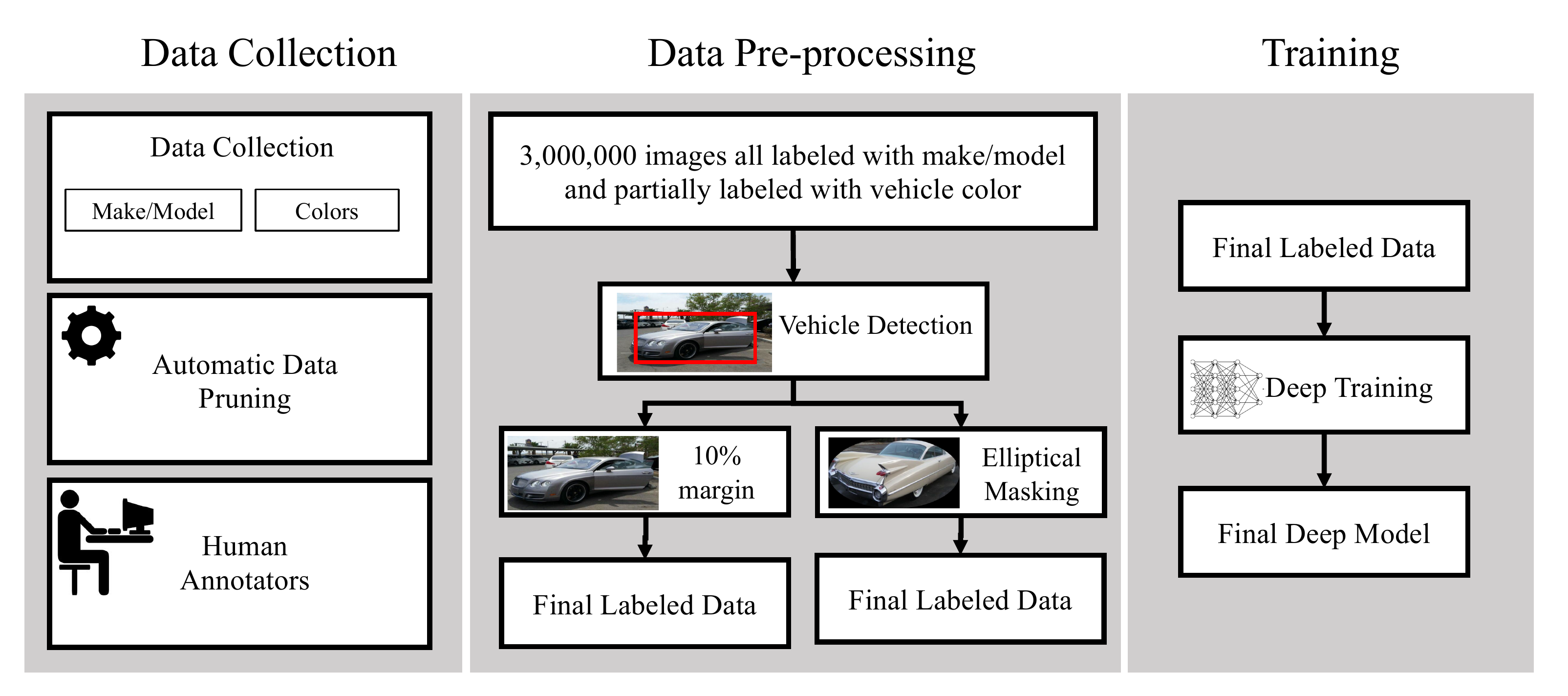}

   \caption{This figure shows the pipeline of our system. Images are collected from different sources. They are later pruned using a semi-automated process with a team of human annotators in the loop. The images are passed through Sighthound's vehicle detector and steps are taken to align them. The images are then fed to our proprietary deep network for training.}
   \label{figPipeline}

\end{center}
\end{figure}

The overview of our system is shown in Figure \ref{figPipeline}. Our training consists of a 3-stage processing pipeline including data collection, data pre-processing and deep training. Data collection plays an important role in our final results, thus collecting data, which requires the least effort in labeling, is of great importance. We collected a large dataset with two different sets of annotations. All the images are annotated with their corresponding vehicle make and model and part of the data is annotated with vehicle colors \footnote{Please note the number of color categories is far less than number of vehicle models}. In order to prepare the final training data we further process the images to eliminate the effect of background. Finally these images are fed into two separate deep neural networks to train the final model. 

\section{Training}

Below we describe in more detail different components of our 3-stage training procedure.

\begin{itemize}
\renewcommand{\labelitemi}{\scriptsize$\blacksquare$} 
\item{\textbf{Data Collection}}:
Data collection plays an important role in training any deep neural network, especially when it comes to fine-grained classification tasks. To address this issue we collected the largest vehicle dataset known to date, where each image is labeled with corresponding make and model of the vehicle. We initially collected over $5$ million images from various sources. We developed a semi-automated process to partially prune the data and remove the undesired images.
We finally used a team of human annotators to remove any remaining errors from the dataset. The final set of data contains over $3$ million images of vehicles with their corresponding make and model tags. Additionally, we labeled part of this data with the corresponding color of the vehicle, chosen from a set of $10$ colors; blue, black, beige, red, white, yellow, orange, purple, green and gray.\\

\item{\textbf{Data Pre-processing}}: An important step in our training is alignment. In order to align images such that all the labeled vehicles are centered in the image, we used Sighthound's vehicle detection model available through the Sighthound Cloud API \footnote{https://www.sighthound.com/products/cloud}. Vehicle detection not only helps us align images based on vehicle bounding boxes but also reduces the impact of the background. This is especially important when there is more than one vehicle in the image. Finally we consider a $10\%$ margin around the vehicle box to compensate for inaccurate (or very tight) bounding boxes. For the task of color recognition, we took pains to further eliminate any influence the background may have on the outcome. To achieve this, we further mask the images with an elliptical mask as shown in Figure \ref{figPipeline}. Note that in certain cases the elliptical mask removes some boundary information of the vehicle. However, this had little effect on the color classification accuracy. \\


\item{\textbf{Deep training}}: The final stage of our pipeline in Figure \ref{figPipeline} involves training two deep neural networks. One is trained to classify vehicles based on their make and model and the other is trained to classify vehicles based on their color. Our networks are designed such that they achieve high accuracy while remaining computationally inexpensive. We trained our networks for four days on four GTX TITAN X PASCAL GPUs. Once the model is trained, we can label images at $150$ fps in batch processing mode.

\end{itemize}

\section{Experiments on SIN $2014$ Test set}
In this section, we report experimental results on two publicly available datasets; the Stanford Cars dataset \cite{KrauseCVPR15} and the Comprehensive Car (compCar) dataset \cite{compCarDataset}. 
The Stanford Cars dataset consists of $196$ classes of cars with a total of $16,185$ images. The data is divided into almost a 50-50 train/test split with $8,144$ training images and $8,041$ testing images. Categories are typically at the level of Make, Model, Year. This means that several categories contain the same model of a make, and the only difference is the year that the car is made. Our original model is not trained to classify vehicle models based on the year of their production. However, after fine-tuning our model on the Stanford Cars training data, we observe that we can achieve better results compared to previously published methods. This is mainly due to the sophistication in the design of our proprietary deep neural network as well as the sizable amount of data used to train this network. The quantitative results are shown in Table \ref{tab:stanford}.

\begin{table}[ht!]
  \caption{Top-1 car classification accuracy on Stanford car dataset.}
  \centering
 \begin{tabular}{ | c | c | }
    \hline
          Methods                                         & Accuracy (top1)   \\ \hline\hline
  \textbf{ \color{red}{Sighthound}}           & \textbf{ \color{red}{$93.6$\%}  }  \\ \hline
   Krause et al. \cite{KrauseCVPR15}      & $92.8$\%   \\ \hline
   Lin et al. \cite{YuICCV2015}                  & $91.3$\%   \\ \hline
   Zhang et al. \cite{ZhangCVPR2016}                  & $88.4$\%   \\ \hline
   Xie et al. \cite{XieCVPR2015}               & $86.3$\%   \\ \hline
   Gosselin et al. \cite{GosselinPR2014}        & $82.7$\%   \\ \hline

    \end{tabular}
\label{tab:stanford}
\end{table}

We also report results on the Comprehensive Car dataset which has recently been published. The task here is to classify data into $431$ different classes based on vehicle make and model. The data is divided into $70\%$ training and $30\%$ testing. There are a total of $36,456$ training images and $15,627$ test images. The top-1 and top-5 accuracy are reported in Table \ref{tab:compCar}. We compare our results with the popular deep network architectures reported in \cite{compCarDataset}. We can clearly see that our fine-tuned model outperforms the existing methods by $4.68\%$ in top-1 accuracy. It is also worth noting that the our model is an order of magnitude faster than GoogLeNet. \\

\begin{table}[ht!]
  \caption{Top-1 and top-5 car classification accuracy of compCar dataset. We compare our results with popular deep networks of GoogLeNet, Overfeat and AlexNet reported in \cite{compCarDataset} }
  \centering
 \begin{tabular}{ | c | c | c |}
    \hline
          Methods                                         & Accuracy (top1)                            & Accuracy (top5)   \\ \hline\hline
  \textbf{ \color{red}{Sighthound}}            & \color{red}\textbf{$95.88$\%}      &\color{red}{\textbf{$99.53$\% }}  \\ \hline
   GoogLeNet \cite{compCarDataset}      & $91.2$\%                  & $98.1$\%                 \\ \hline
   Overfeat \cite{compCarDataset}           & $87.9$\%                  & $96.9$\%                  \\ \hline
   AlexNet \cite{compCarDataset}             & $81.9$\%                 & $94.0$\%                  \\ \hline

    \end{tabular}
\label{tab:compCar}
\end{table}

Lastly we test the verification accuracy of the proposed method on compCar dataset. The compCar dataset includes three sets of data for verification experiments, sorted by their difficulties. Each set contains 20,000 pairs of images. The likelihood ratio of each image pair is obtained by  computing the euclidean distant between features computed using our the deep network. The likelihood ratio is then compared against a threshold to make the final decision. The results are shown in Table \ref{tab:verification}. As can be seen our model, fine-tuned on the verification training data of compCar dataset, outperforms other methods. It is worth to mention that , even without fine-tuning our features can achieve a high verification accuracy of $92.03\%$, $86.52\%$, $80.17\%$ on different sets of easy, medium and hard respectively. 

\begin{table}[ht!]
  \caption{Verification accuracy of three different sets, easy, medium and hard in \cite{compCarDataset}. Sighthound is our fine-tuned model trained on Sighthound data. It can be seen that the model outperforms previous methods by a large margin.}
  \centering
 \begin{tabular}{ | c | c | c | c |}
    \hline
          Methods                          & Accuracy(Easy)     & Accuracy(Medium)  & Accuracy(Hard)   \\ \hline\hline
  \textbf{ \color{red}{Sighthound}}   & \color{red}\textbf{$93.00$\%}  &\color{red}{\textbf{$86.18$\% }} &\color{red}{\textbf{$80.05$\% }}  \\ \hline
    
   Yang et. al. \cite{compCarDataset}      & $83.3$\%        & $82.4$\%      & $76.1$\%           \\ \hline
   Sochor et. al. \cite{SochorCVPR16}     & $85.0$\%        & $82.7$\%      & $76.8$\%            \\ \hline

    \end{tabular}
\label{tab:verification}
\end{table}

\section{Quantitative Results}

\begin{figure}[!ht]
\begin{center}
   \includegraphics[width=\linewidth]{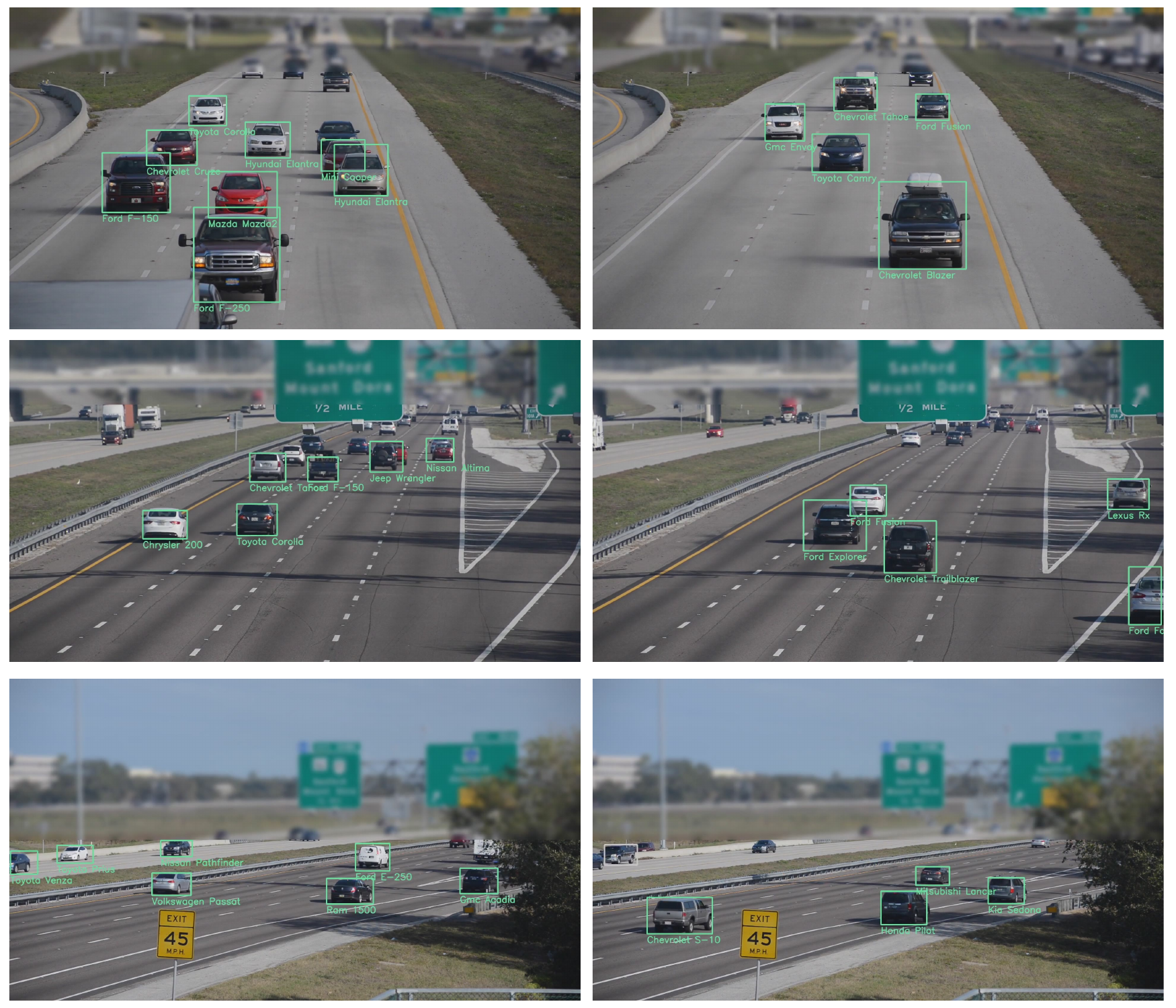}

   \caption{Quantitative results of the proposed method in surveillance cameras.}
   \label{fig:quant1}

\end{center}
\end{figure}

\begin{figure}[!ht]
\begin{center}
   \includegraphics[width=\linewidth]{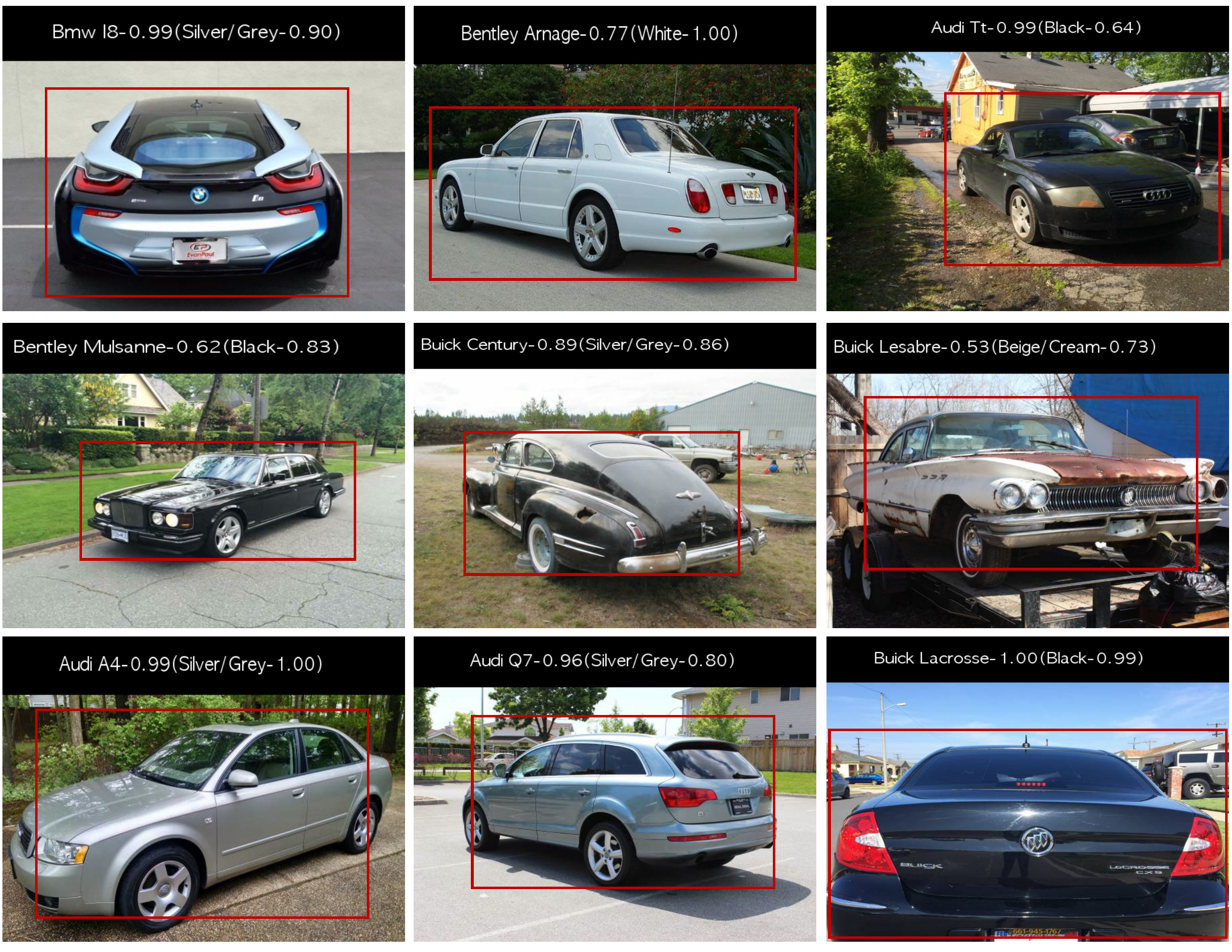}

   \caption{Quantitative results of the proposed method.}
   \label{fig:quant2}

\end{center}
\end{figure}

We demonstrate some quantitative results in Figures \ref{fig:quant1} and \ref{fig:quant2}, capturing different scenarios. Figure \ref{fig:quant1} shows results for images mostly taken by people. Figure \ref{fig:quant2} shows a surveillance-like scenario where the camera is mounted at a higher distance from the ground. These images are illustrative of the robustness of our large training dataset, captured from different sources, to real world scenarios. 

\section{Conclusions}

In this paper we presented an end to end system for vehicle make, model and color recognition. The combination of Sighthound's novel approach to the design and implementation of deep neural networks and a sizable dataset for training allow us to label vehicles in real time with high degrees of accuracy.We conducted several experiments for both classification and verification tasks on public benchmarks and showed significant improvement over previous methods.

\newpage
\bibliographystyle{splncs}
\bibliography{egbib}

\end{document}